\documentclass[12pt]{article}
\usepackage[margin=1in]{geometry}

\usepackage{natbib}
\usepackage[utf8]{inputenc} % allow utf-8 input
\usepackage[T1]{fontenc}    % use 8-bit T1 fonts
\usepackage{hyperref}       % hyperlinks
\usepackage{url}            % simple URL typesetting
\usepackage{booktabs}       % professional-quality tables
\usepackage{amsfonts}       % blackboard math symbols
\usepackage{nicefrac}       % compact symbols for 1/2, etc.
\usepackage{microtype}      % microtypography

\usepackage{morenotations}

\title{\papertitle}
\author{Hisham Husain \\ 
        The Australian National University \& Data61\\
        \texttt{hisham.husain@anu.edu.au}}
\begin{document}

\date{}

\maketitle

\begin{abstract}
Robustness to adversarial attacks is an important concern due to the fragility of deep neural networks to small perturbations and has received an abundance of attention in recent years. Distributionally Robust Optimization (DRO), a particularly promising way of addressing this challenge, studies robustness via divergence-based uncertainty sets and has provided valuable insights into robustification strategies such as regularization. In the context of machine learning, the majority of existing results have chosen $f$-divergences, Wasserstein distances and more recently, the Maximum Mean Discrepancy (MMD) to construct uncertainty sets. We extend this line of work for the purposes of understanding robustness via regularization by studying uncertainty sets constructed with Integral Probability Metrics (IPMs) - a large family of divergences including the MMD, Total Variation and Wasserstein distances. Our main result shows that DRO under \textit{any} choice of IPM corresponds to a family of regularization penalties, which recover and improve upon existing results in the setting of MMD and Wasserstein distances. Due to the generality of our result, we show that other choices of IPMs correspond to other commonly used penalties in machine learning. Furthermore, we extend our results to shed light on adversarial generative modelling via $f$-GANs, constituting the first study of distributional robustness for the $f$-GAN objective. Our results unveil the inductive properties of the discriminator set with regards to robustness, allowing us to give positive comments for several penalty-based GAN methods such as Wasserstein-, MMD- and Sobolev-GANs. In summary, our results intimately link GANs to distributional robustness, extend previous results on DRO and contribute to our understanding of the link between regularization and robustness at large.
\end{abstract}

\newpage

\section{Introduction}
Robustness to adversarial attacks is an important concern due to the fragility of deep neural networks to small perturbations and has received an abundance of attention in recent years \citep{goodfellow2014explaining, szegedy2013intriguing, madry2017towards}. Distributionally Robust Optimization (DRO), a particularly promising way of addressing this challenge, studies robustness via divergence-based uncertainty sets and considers robustness against shifts in distributions. To see this more clearly, for some space $\Omega$, model $h: \Omega \to \mathbb{R}$ and training data $\hat{P}$ with empirical loss $\E_{x \sim \hat{P}}[l_f]$, DRO studies the objective $\sup_{Q \in \mathcal{U}} \E_{x \sim Q}[l_f]$ where $\mathcal{U} = \braces{Q : d(Q,\hat{P}) \leq \epsilon}$ for a given divergence $d$ and $\epsilon > 0$ that characterize the adversary. Work along this line has shown that this objective is upper bounded by the empirical loss $\E_{x \sim \hat{P}}[l_f]$ plus a penalty term that plays the role of a regularizer, consequently providing formal connections and valuable insights into regularization as a robustification strategy \citep{gotoh2018robust,lam2016robust,namkoong2017variance, ben2013robust,duchi2013local,cranko2020generalised}. 

The choice of $d$ is crucial as it highlights the strength and nature of robustness we desire, and different choices yield differing penalties. It has been shown that minimizing the distributionally robust objective when $d$ is chosen to be an $f$-divergence is roughly equivalent to variance regularization \citep{gotoh2018robust,lam2016robust,namkoong2017variance}. However, there is a problem with this choice of $d$, as highlighted in \citep{staib2019distributionally}: every distribution in the uncertainty set is required to be absolutely continuous with respect to $P$. This is particularly problematic in the case when $P$ is empirical since every distribution in $\mathcal{U}$ will be finitely supported, meaning that the population distribution will not be contained as it is typically continuous.

Choosing the Wasserstein distance as $d$ is a typical antidote for this problem, and much work has been invested in this direction, explicating connections to Lipschitz regularization \citep{gao2016distributionally, cisse2017parseval, sinha2017certifiable, shafieezadeh2019regularization, cranko2020generalised}. More recently, uncertainty sets based on the kernel Maximum Mean Discrepancy (MMD) were investigated to address concerns with the $f$-divergence and discovered links to regularization with Hilbert space norms. Both the Wasserstein distance and MMD are part of a larger family of divergences referred to as Integral Probability Metrics (IPM) \citep{muller1997integral}, which are characterized by a set of functions $\mathcal{F}$, and include other metrics such as the Total Variation distance and the Dudley Metric \citep{sriperumbudur2009integral}.  

%More recently, another line of work with intentions of addressing drawbacks of $f$-divergences, considered uncertainty sets derived the kernel Maximum Mean Discrepancy (MMD), which like the Wasserstein distance is a moment-matching divergence \citep{mohamed2016learning}. It was then shown that distributional robustness in this setting roughly amounts to regularizing the Hilbert space norm. Both these divergences, however, are part of a larger family of distribution, known as the Integral Probability Metric (IPM) \citep{muller1997integral} which are characterized by a set of functions $\mathcal{F}$, and include other metrics such as the Total Variation and Dudley Metric \citep{sriperumbudur2009integral}.
In this work, we generalize these results and study DRO for uncertainty sets induced by the Integral Probability Metric (IPM) for \textit{any} set of functions $\mathcal{F}$. We present an identity which links distributional robustness under these uncertainty sets $\mathcal{U}_{\mathcal{F}}$, to regularization under a new penalty $\Lambda_{\mathcal{F}}$. Our identity takes the form
\begin{align}\label{intro:eq1}
    \boxed{\sup_{Q \in \mathcal{U}_{\mathcal{F}}} \int_{\Omega} h dQ = \int_{\Omega} h dP + \Lambda_{\mathcal{F}}(h)}
\end{align}
The appeal of this result is that it reduces the infinite-dimensional optimization on the left-hand side into a penalty-based regularization problem on the right-hand side. We study properties of this penalty and show that it can be upper bounded by another term, $\Theta_{\mathcal{F}}$, which recovers and improves upon existing penalties when $\mathcal{F}$ is chosen to coincide with the MMD and Wasserstein distances. Our result, however, holds in much more generality, allowing us to derive new penalties by considering other IPMs such as the Total Variation, Fisher IPM \citep{mroueh2017fisher}, and Sobelov IPM \citep{mroueh2017sobolev}. We find that these new penalties are related to existing penalties in regularized critic losses \citep{thanh2019improving} and manifold regularization \citep{belkin2006manifold}, permitting us to provide untried robustness perspectives for existing regularization schemes. Furthermore, most work in this direction takes the form of upper bounds, and although working with $\Theta_{\mathcal{F}}$ reduces (\ref{intro:eq1}) into an inequality, we present a necessary and sufficient condition such that $\Lambda_{\mathcal{F}}$ coincides with $\Theta_{\mathcal{F}}$, yielding equality. This condition reveals an intimate connection between distributional robustness and regularized binary classification. 

We then apply our result to understanding the distributional robustness of Generative Adversarial Networks (GANs), a popular method for modelling distributions that learn a model $Q$ by utilizing a set of discriminators $D$ that try to distinguish $Q$ from $P$ (the training data). Our analysis applies to the $f$-GAN objective \citep{nowozin2016f} - a loss that subsumes many existing GAN losses. This is, to the best of our knowledge, the first analysis of robustness for $f$-GANs with respect to divergence-based uncertainty sets. An investigation into the robustness of GANs is of topical interest \cite[Problem~7]{odena2019open} since GANs use discriminator and adversarial based objectives to drive learning, which suggests there is a natural application to use them to train robust classifiers \citep{wang2019direct,charlier2019syngan,zhao2017generating,zhao2019perturbations,lee2017generative,jalal2017robust,poursaeed2018generative,song2017pixeldefend,song2018constructing,hayes2018learning,xiao2018generating,samangouei2018defense}. Our result tells us that the model learned by a GAN is robust depending on the complexity of discriminators $D$, forming a discrimination-robustification trade-off which parallels and extends previous discrimination-generalization trade-offs \citep{zhang2017discrimination}. Our result also complements existing results that link discriminator complexity to the stability of training \citep{farnia2018convex,liu2018inductive,zhou2019lipschitz}. Furthermore, our findings allow us to give positive results and robustness perspectives for many existing methods that use restricted discriminator sets such as MMD-GAN \citep{li2017mmd,arbel2018gradient, binkowski2018demystifying}, Wasserstein-GAN \citep{arjovsky2017wasserstein, gulrajani2017improved}, Sobelov-GAN \citep{mroueh2017sobolev}, Fisher-GAN \citep{mroueh2017fisher} and other penalty-based GANs \citep{thanh2019improving}.

Our contributions come in three Theorems, where the first two concern DRO with IPMs (Section 3) and the third is an extension to understanding GANs (Section 4):\\
\noindent $\triangleright$ (Theorem \ref{main-ipm-robustness-thm}) An identity for distributional robustness using uncertainty sets induced by any IPM. Our result tells us that this is \textit{exactly} equal to regularization with a penalty $\Lambda_{\mathcal{F}}$. We show that this penalty can be upper bounded by another penalty $\Theta_{\mathcal{F}}$ which recovers existing work when the IPM is set to the MMD and Wasserstein distance, tightening these results. Since our result holds in much more generality, we derive penalties for other IPMs such as the Total Variation, Fisher IPM, and Sobelov IPM, and draw connections to existing methods. \\
\noindent $\triangleright$ (Theorem \ref{nec-suff-condition}) A necessary and sufficient condition under which the penalties $\Lambda_{\mathcal{F}}$ and $\Theta_{\mathcal{F}}$ coincide. It turns out this condition is linked to regularized binary classification and is related to critic losses appearing in penalty-based GANs. This allows us to give positive results for work in this direction, along with drawing a link between regularized binary classification and distributional robustness. \\
\noindent $\triangleright$ (Theorem \ref{GAN-robust-thm}) A result that characterizes the distributional robustness of the $f$-GAN objective showing that the discriminator set plays an important part for the robustness of a GAN. This is, to the best of our knowledge, the first result on divergence-based distributional robustness of $f$-GANs. Our result allows us to provide a novel perspective for several existing penalty-based GAN methods such as Wasserstein-, MMD-, and Sobelov-GANs.
\section{Preliminaries}
\subsection{Notation}
We will use $\Omega$ to denote a Polish space and denote $\Sigma$ as the standard Borel $\sigma$-algebra on $\Omega$ and $\mathbb{R}$ will denote the real numbers. We use $\mathscr{F}(\Omega, \mathbb{R})$ to denote the set of all bounded and measurable functions mapping from $\Omega$ into $\mathbb{R}$ with respect to $\Sigma$, $\mathscr{B}(\Omega)$ to be the set of finite signed measures and the set $\mathscr{P}(\Omega) \subset \mathscr{B}(\Omega)$ will denote the set of probability measures. For any additive monoid $X$, a function $f:X \to \mathbb{R}$ is subadditive if $f(x+x') \leq f(x) + f(x')$ and the \textit{infimal convolution} between two functions $f: X \to \mathbb{R}$ and $g: X \to \mathbb{R}$ is another function given by $(f \iconv g) (x) = \inf_{x' \in X} \bracket{ f(x') + g(x-x') }$. For any proposition $\mathscr{I}$, the inversion bracket is $\llbracket \mathscr{I} \rrbracket = 1$ if $\mathscr{I}$ is true and $0$ otherwise. We say a set of functions $\mathcal{F}$ is even if $h \in \mathcal{F}$ implies $-h \in \mathcal{F}$. For a function $h \in \mathscr{F}(\Omega, \mathbb{R})$ and metric $c: \Omega \times \Omega \to \mathbb{R}$, the Lipschitz constant of $h$ (w.r.t $c$) is $\operatorname{Lip}_c(h) = \sup_{\omega, \omega' \in \Omega} \card{h(\omega) - h(\omega')}/c(\omega, \omega')$ and $\nrm{h}_{\infty} := \sup_{\omega \in \Omega} \card{h(\omega)}$. For any set of functions $\mathcal{F} \subseteq \mathscr{F}(\Omega, \mathbb{R})$, we use $\chull{\mathcal{F}}$ to denote the closed convex hull of $\mathcal{F}$. For a function $h \in \mathscr{F}(\Omega, \mathbb{R})$ and measure $\mu \in \mathscr{P}(\Omega)$, we use $\operatorname{Var}_{\mu}(h) = \E_{\mu}[h^2] - \E_{\mu}[h]^2$ to denote the variance of $h$ under $\mu$.

\subsection{Background and Related Work}
We will focus our discussion around Distributionally Robust Optimization (DRO) \citep{scarf1957min} and its use for understanding machine learning. For a given reference distribution $P$, which is typically the training data in machine learning, the neighbourhood takes the form $\braces{Q : d(Q,P) \leq \epsilon}$ for some divergence $d$ and $\epsilon > 0$ that characterize the nature and budget of robustness. In the context of machine learning, the most popular choices of $d$ studied thus far are the $f$-divergences \citep{ben2013robust, duchi2016statistics, lam2016robust}, Wasserstein distance \citep{esfahani2018data, abadeh2015distributionally, blanchet2019robust} and the kernel Maximum Mean Discrepancy (MMD) \citep{staib2019distributionally}. For two distributions $P,Q$, the $f$-divergence is $d_f(P,Q) = \int_{\Omega} f(dP/dQ) dQ$ and the main advancement regarding $f$-divergences, centered around $\chi^2$-divergence, is the connection to variance regularization \citep{gotoh2018robust,lam2016robust,namkoong2017variance}. This is appealing since it reflects the classical bias-variance trade-off. In contrast, variance regularization also appears in our results, under the choice of $\mu$-Fisher IPM. One of the drawbacks of using $f$-divergences as pointed out in \citep{staib2019distributionally}, is that the uncertainty set induced by $f$-divergences contains only those distributions that share support (since we require absolute continuity) and thus will typically not include the population distribution. The Wasserstein distance is commonly antidotal for these problems since it is defined between distributions that do not share support and DRO results have been developed for this direction, with the main results showing links to Lipschitz regularization \citep{gao2016distributionally, cisse2017parseval, sinha2017certifiable, shafieezadeh2019regularization, cranko2020generalised}. Another distance used to remedy this problem is the Maximum Mean Discrepancy, which has been studied in \citep{staib2019distributionally} and shown connections to Hilbert space norm regularization and kernel ridge regression. Since both of these are Integral Probability Metrics (IPMs) \citep{muller1997integral}, it is natural to study uncertainty sets generated by general IPMs:
\begin{definition}[Integral Probability Metric]
For any $\mathcal{F} \subseteq \mathscr{F}(\Omega, \mathbb{R})$, the ($\mathcal{F}$-)Integral Probability Metric between $P,Q \in \mathscr{P}(\Omega)$ is
\begin{align*}
    d_{\mathcal{F}}(P,Q) := \sup_{h \in \mathcal{F}} \bracket{ \int_{\Omega}h dP - \int_{\Omega} h dQ }.
\end{align*}
\end{definition}
The IPM is characterized by a set $\mathcal{F}$ and if $\mathcal{F}$ is even, then $d_{\mathcal{F}}$ is symmetric. One should note that we have an intersection between IPMs and $f$-divergence when $\mathcal{F} = \braces{h : \nrm{h}_{\infty} \leq 1}$ and $f(t) = \card{t-1}$, which corresponds to the Total Variation. Other cases when they intersect have been thoroughly pursued in \citep{sriperumbudur2009integral}. Another interesting case is the 1-Wasserstein distance, which is realized when $\mathcal{F} = \braces{h : \operatorname{Lip}_c(h) \leq 1}$ for some ground metric $c: \Omega \times \Omega \to \mathbb{R}$ \citep{vOT}. Table \ref{ipm-examples} contains other known choices of IPMs. As the IPM can be viewed as matching moments specified by $\mathcal{F}$, there is similar work which considers uncertainty sets that match the first and second moment such as \citep{delage2010distributionally}. In the context of machine learning our work is, to the best of our knowledge, the first study of the general IPM to understand regularization. Outside this realm, there exist pursuits to study structural properties of IPM-based uncertainty sets such as invariance \citep{shapiro2017distributionally}. While these are important to understand, they, however, do not give immediate consequences for machine learning.
\section{Distributional Robustness}
\begin{table}[]
\centering
\def\arraystretch{2}
\caption{}
\begin{tabular}{lll}
\hline
IPM                      & $\mathcal{F}$                                                        & $\Theta_{\mathcal{F}}(h)$                              \\ \hline
Wasserstein Distance     & $\braces{h : \operatorname{Lip}_c(h) \leq 1}$                        & $\operatorname{Lip}_c(h)$                              \\
Maximum Mean Discrepancy & $\braces{h :  \nrm{h}_{k} \leq 1}$                                   & $\nrm{h}_k$                                            \\
Total Variation          & $\braces{h :  \nrm{h}_{\infty} \leq 1}$                              & $\nrm{h}_{\infty}$                                     \\
Dudley Metric            & $\braces{h : \nrm{h}_{\infty} + \operatorname{Lip}_c(h) \leq 1}$     & $\nrm{h}_{\infty} + \operatorname{Lip}_c(h)$           \\
$\mu$-Sobelov IPM        & $\braces{h :  \E_{\mu(X)}\left[ \nrm{\nabla h(x)}^2 \right] \leq 1}$ & $\sqrt{\E_{\mu(X)}\left[ \nrm{\nabla h(x)}^2 \right]}$ \\
$\mu$-Fisher IPM         & $\braces{h : \E_{\mu(X)}\left[ h^2(X) \right]  \leq 1}$              & $\sqrt{\E_{\mu(X)}\left[ h^2(X) \right]}$              \\ \hline
\end{tabular}
\label{ipm-examples}
\end{table}
\begin{wrapfigure}{R}{0.4\textwidth}
    \centering
    \vspace{-10pt}
    \includegraphics[width=0.4\textwidth,height=0.36902173913\textwidth]{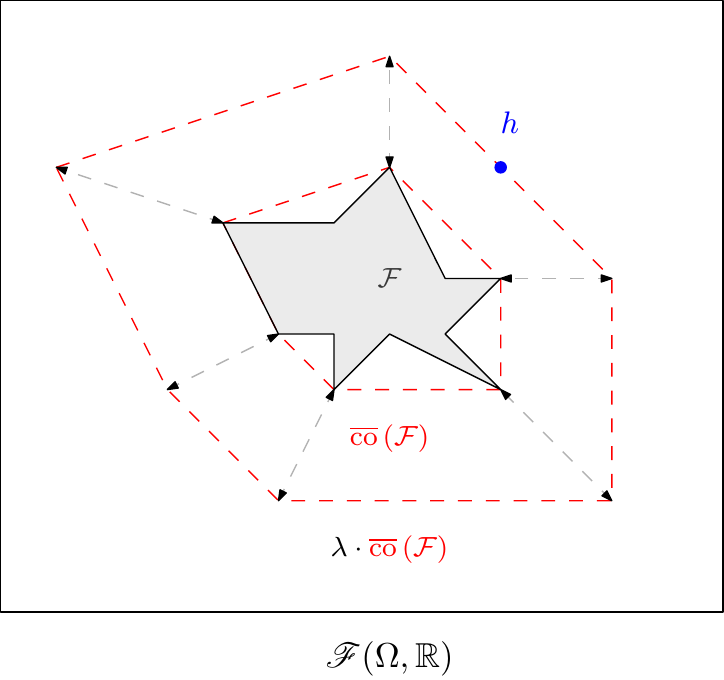}
    \vspace{-15pt}
    \caption{$\Theta_{\mathcal{F}}(h)$ is the smallest multiplicative factor $\lambda$ required to stretch the convex hull of $\mathcal{F}$ until $h$ is contained.}
    \label{fig:thetaF}
    \vspace{-40pt}
\end{wrapfigure}
In this section, we first introduce the uncertainty set and two complexity measures that form building blocks of the main penalty term $\Lambda_{\mathcal{F}}$ (as appearing in Equation \ref{intro:eq1}), then proceed to the main distributional robustness Theorem.
\begin{definition}
For any $\mathcal{F} \subseteq \mathscr{F}(\Omega, \mathbb{R})$, $P \in \mathscr{P}(\Omega)$, the $\mathcal{F}$-ball centered at $P$ with radius $\epsilon$ is defined to be $B_{\epsilon, \mathcal{F}}(P) = \braces{Q \in \mathscr{P}(\Omega) : d_{\mathcal{F}}(Q,P) \leq \epsilon }$.
\end{definition}
We now introduce a complexity measure that will be of central importance when defining the penalty: For a function set $\mathcal{F} \subseteq \mathscr{F}(\Omega, \mathbb{R})$ and function $h \in \mathscr{F}(\Omega, \mathbb{R})$, we set $\Theta_{\mathcal{F}} (h) := \inf\braces{\lambda > 0 : h \in \lambda \cdot \chull{ \mathcal{F}} }$. This quantity represents the smallest lambda that multiplicatively stretches the set $\chull{\mathcal{F}}$ until it contains $h$. We illustrate this geometrically in Figure \ref{fig:thetaF} for a non-convex case of $\mathcal{F}$ and present examples of $\Theta_{\mathcal{F}}$ in Table \ref{ipm-examples}.

The second complexity measure depends on a distribution $P \in \mathscr{P}(\Omega, \mathbb{R})$ and is defined as $J_P(h) = \sup_{\nu \in \mathscr{P}(\Omega)} \int_{\Omega} h d\nu - \int_{\Omega} h dP$. Note that if $h$ reaches its maximum at some $\omega^{*} \in \Omega$ then $J_P(h)$ will be smaller if $P$ is concentrated around $\omega^{*}$. We now present the main penalty, which is \text{infimal convolution} of these two complexity measures.
\begin{definition}[$\mathcal{F}$-Penalty]
For any $\mathcal{F} \subseteq \mathscr{F}(\Omega, \mathbb{R})$, $h \in \mathscr{F}(\Omega, \mathbb{R})$ and $\epsilon > 0$, the $\mathcal{F}$-penalty $\Lambda_{\mathcal{F},\epsilon}: \mathscr{F}(\Omega, \mathbb{R}) \to [0, \infty]$ is
\begin{align*}
    \Lambda_{\mathcal{F},\epsilon}(h) = \bracket{J_P \iconv \epsilon \Theta_{\mathcal{F}}}(h),
\end{align*}
where $J_P(h) = \sup_{\nu \in \mathscr{P}(\Omega)} \int_{\Omega} h d\nu - \int_{\Omega} h dP$ and $\iconv$ is the infimal convolution operator.
\end{definition}
The infimal convolution is central in convex analysis since it is the analogue of addition in the convex dual space \citep{stromberg1994study}. We now present the main theorem, which links this penalty to distributional robustness via $\mathcal{F}$-uncertainty sets and discuss further the role of this penalty.
\begin{theorem}
\label{main-ipm-robustness-thm}
Let $\mathcal{F} \subseteq \mathscr{F}(\Omega, \mathbb{R})$ and $P \in \mathscr{P}(\Omega)$. For any $h \in \mathscr{F}(\Omega, \mathbb{R})$ and for all $\epsilon > 0$
\begin{align*}
    \sup_{Q \in B_{\epsilon, \mathcal{F}}(P)} \int_{\Omega} h dQ = \int_{\Omega} h dP + \Lambda_{\mathcal{F}, \epsilon}(h).
\end{align*}
\end{theorem}
\begin{proof}(Sketch, full proof in the Appendix) We can rewrite the constraint over $B_{\epsilon, \mathcal{F}}(P)$ with the use of a dual variable which leads to a min-max equation. Using generalized minimax theorems \citep{fan1953minimax} and compactness of the set of probability measures, we are able to swap the min-max and solve the inner min using classical results in convex analysis \citep{penot2012calculus}, yielding the statement of the theorem.
\end{proof}
The result allows us to turn the infinite-dimensional optimization on the left-hand side into a familiar penalty-based regularization objective, and we remark that there is no restriction on the choice of $\mathcal{F}$. To see the effect of $\Lambda_{\mathcal{F}, \epsilon}$, notice that by definition of $\iconv$ we have
\begin{align*}
    \Lambda_{\mathcal{F},\epsilon}(h) =  \inf_{\substack{h_1, h_2\\ h_1 + h_2 = h}} \bracket{J_P(h_1) + \epsilon \Theta_{\mathcal{F}}(h_2)},
\end{align*}
which means this penalty finds a decomposition of $h$ into $h_1,h_2$ so that the two penalties $J_P(h_1)$ and $\epsilon \Theta_{\mathcal{F}}(h_2)$ are controlled. Notice that any decomposition gives an upper bound, and this is precisely how we will show links and tighten existing results. We will then present a necessary and sufficient condition under which $\Lambda_{\mathcal{F}, \epsilon}(h) = \epsilon \Theta_{\mathcal{F}}(h)$. This condition plays a fundamental role in linking robustness to regularization and unlike majority of existing results, yields an \textit{equality}. 

To see the applicability of the result, consider the supervised learning setup: We have an input space $\mathcal{X}$, output space $\mathcal{Y}$, and a loss function $l: \mathcal{Y} \times \mathcal{Y} \to \mathbb{R}$ which measures performance of a hypothesis $g: \mathcal{X} \to \mathcal{Y}$ on a sample $(x,y)$ with $l(g(x),y)$. In this case, we set $\Omega = \mathcal{X} \times \mathcal{Y}$, $P$ to be the available data, and $h = l(g(x),y)$:
\begin{align*}
    \sup_{Q \in B_{\epsilon, \mathcal{F}}(P)} \int_{\Omega} l(g(x),y) dQ(x,y) = \underbrace{\int_{\Omega} l(g(x),y) dP(x,y) }_{\mathclap{\text{data fitting term}}} + \underbrace{\Lambda_{\mathcal{F}, \epsilon}(l(g(x),y))}_{\mathclap{\text{robustness penalty}}}.
\end{align*}
The first term is interpreted as a data fitting term, while the second term is a penalty term that ensures robustness of $g$. We remark that upper bounds are still favourable in the application of supervised learning, which we will now discuss.

To generate our first upper bound, consider the following decomposition: $h_1 = b$ and $h_2 = h - b$ for some $b \in \mathbb{R}$, yielding the following Corollary.
\begin{corollary}
\label{theta-upperbound}
Let $\mathcal{F} \subseteq \mathscr{F}(\Omega, \mathbb{R})$ and $P \in \mathscr{P}(\Omega)$. For any $h \in \mathscr{F}(\Omega, \mathbb{R})$ and for all $\epsilon > 0$
\begin{align*}
\sup_{Q \in B_{\epsilon, \mathcal{F}}(P)} \int_{\Omega} h dQ \leq \int_{\Omega} h dP + \epsilon \inf_{b \in \mathbb{R}}\Theta_{\mathcal{F}}(h - b).
\end{align*}
\end{corollary}
We will show that Corollary \ref{theta-upperbound} recovers or tightens main results, and holds in much more generality since we may choose \textit{any} set $\mathcal{F}$. The choice of $\mathcal{F}$ is important to our notion of uncertainty as it captures the moments we are interested in, and there is a natural trade-off between picking $\mathcal{F}$ to be too large or too small, which we illustrate with extreme cases. Consider the largest possible set $\mathcal{F} = \mathscr{F}(\Omega, \mathbb{R})$, under which the uncertainty set of distributions, $B_{\epsilon, \mathcal{F}}(P) = \braces{P}$ is a singleton for all $\epsilon > 0$. This is indeed reflected on the right hand side of Corollary \ref{theta-upperbound}, noting that such a strong set $\mathcal{F}$ yields $\Theta_{\mathcal{F}}(h) = 0$ for any $h \in \mathscr{F}(\Omega, \mathbb{R})$. On the other hand, if we pick $\mathcal{F} = \braces{f(x) = k  : k \in \mathbb{R}}$ to be the set of constants, which is a rather restrictive set, then the uncertainty ball of distributions is the largest it can be $B_{\epsilon, \mathcal{F}} = \mathscr{P}(\Omega)$ since $d_{\mathcal{F}}(Q,P) = 0$ for all $Q \in \mathscr{P}(\Omega)$. We now focus on non-trivial settings of $\mathcal{F}$, showing that $\Theta_{\mathcal{F}}$ recovers and improves upon familiar existing penalties.
\begin{enumerate}[(a),itemindent=*,leftmargin=*]
\item \textbf{(Wasserstein Distance)} $\mathcal{F} = \braces{h : \operatorname{Lip}_c(h) \leq 1}$. The penalty is $\Theta_{\mathcal{F}}(h) = \operatorname{Lip}_c(h)$, and Corollary \ref{theta-upperbound} recovers the intuition of Lipschitz regularized networks as presented in \citep{gao2016distributionally, cisse2017parseval, sinha2017certifiable, shafieezadeh2019regularization, cranko2020generalised}. However, the penalty in the original theorem $\Lambda_{\mathcal{F},\epsilon}$ is tighter. To see this by example, consider $\Omega = \mathbb{R}$, $P$ a normal distribution centered at $0$ with variance $\sigma > 0$, $h(t) = \sin 2t + t$ and $\epsilon = 1$. Note that $\epsilon \operatorname{Lip}_c(h) = 3$ however $h$ can be decomposed into $h_1 = \sin 2t$ and $h_2 = t$ with $J_P(h_1) = 1$ and $\epsilon \operatorname{Lip}_c(h_2) = 1$. Hence we have $\Lambda_{\mathcal{F}, \epsilon}(h) \leq 2 < 3 = \epsilon \operatorname{Lip}_c(h)$.
\item \textbf{(Maximum Mean Discrepancy)} $\mathcal{F} = \braces{h : \nrm{h}_k \leq 1}$ where $k: \Omega \times \Omega \to \mathbb{R}$ is a positive definite characteristic kernel and $\nrm{\cdot}_k$ is the Reproducing Kernel Hilbert Space (RKHS) norm induced by $k$ \citep{muandet2016kernel}. For $h$ in the RKHS, the penalty can be bounded by $\Lambda_{\mathcal{F}, \epsilon}(h) \leq \inf_{b \in \mathbb{R}} \nrm{h - b}_k$. This tightens the existing work on MMD DRO \cite[Corollary~3.2]{staib2019distributionally} when $b = 0$.
\item \textbf{(Total Variation)} $\mathcal{F} = \braces{h : \nrm{h}_{\infty} \leq 1}$. Our result tells us that the penalty upper bounded with $\Lambda_{\mathcal{F}, \epsilon}(h) \leq \inf_{b \in \mathbb{R}} \nrm{h - b}_{\infty}$, which is tighter than taking $\nrm{h}_{\infty}$.
\item \textbf{($\mu$-Fisher IPM)} $\mathcal{F} = \braces{h : \E_{\mu(X)}\left[ h^2(X) \right]  \leq 1}$ for some $\mu \in \mathscr{P}(\Omega)$ \citep{mroueh2017fisher}. The penalty is $\Theta_{\mathcal{F}}(h) = \sqrt{\E_{\mu(X)}\left[ h^2(X) \right]}$, however we can solve the infimum in Corollary \ref{theta-upperbound} to get $\inf_{b \in \mathbb{R}} \Theta_{\mathcal{F}}(h-b) = \sqrt{\operatorname{Var}_{\mu}(h) }$ (Lemma \ref{mu-ipm-der} in Supplementary). This is interesting since the variance of $h$ as a penalty has appeared in work studying $f$-divergence uncertainty sets. Note that when $\mu = (P+Q) / 2$ for some $P,Q \in \mathscr{P}(\Omega)$ then $d_{\mathcal{F}}(P,Q)$ is the $\chi^2$-divergence, the central $f$-divergence in these lines of work. In this setting, Corollary \ref{theta-upperbound} extends the interpretation of variance regularization as a robustification strategy for any $\mu \in \mathscr{P}(\Omega)$.
\end{enumerate}
Another interesting choice of $\mathcal{F}$ is the $\mu$-Sobelov IPM which we show in Table \ref{ipm-examples}, whereby the resulting penalty is similar to those existing in manifold regularization \citep{belkin2006manifold}. All IPMs considered so far are of the form $\braces{h : \zeta(h) \leq 1}$ for some $\zeta: \mathscr{F}(\Omega, \mathbb{R}) \to [0,\infty]$, and the resulting $\Theta_{\mathcal{F}}(h)$ closely resembles $\zeta(h)$. We derive $\Theta_{\mathcal{F}}$ for this general form with some assumptions on $\zeta$.
\begin{lemma}
\label{hom-penalties}
Let $\zeta: \mathscr{F}(\Omega, \mathbb{R}) \to [0,\infty]$ be such that for some $k > 0$, $\zeta(a \cdot h) = a^k \cdot \zeta(h)$ for any $h \in \mathscr{F}(\Omega, \mathbb{R}), a> 0$. If $\mathcal{F} = \braces{h : \zeta(h) \leq 1}$, then $\Theta_{\mathcal{F}}(h) \leq \sqrt[k]{\zeta(h)}$ with equality if $\zeta$ is convex.
\end{lemma}
Our examples presented in Table \ref{ipm-examples} have convex choices of $\zeta$ with either $k=1$ or $k=2$. Using this Lemma, we may also interpret the case of two penalties added together, such as the Dudley metric in Table \ref{ipm-examples}. Furthermore, Lemma \ref{hom-penalties} can be used for future applications of our work to elucidate robustness perspectives of methods using penalties of the form $\sqrt[k]{\zeta(h)}$.

We now return to the discussion on how closely related $\Lambda_{\mathcal{F}, \epsilon}$ is to $\epsilon \Theta_{\mathcal{F}}$. Consider now two decompsitions of $h$ for the infimal convolution: $h_1 = 0, h_2 = h$ and $h_1 = h,h_2 = 0$, so we have $\Lambda_{\mathcal{F},\epsilon}(h) \leq \epsilon \Theta_{\mathcal{F}}(h)$ and $\Lambda_{\mathcal{F},\epsilon}(h) \leq J_P(h)$ respectively. This yields $\Lambda_{\mathcal{F},\epsilon}(h) \leq \min\bracket{J_P(h),\epsilon \Theta_{\mathcal{F}}(h)}$, and we illustrate the tightness of this inequality through the following lemma.
\begin{lemma}
\label{penalty-tightness}
The mapping $h \mapsto \Lambda_{\mathcal{F},\epsilon}(h)$ is subadditive and $\Lambda_{\mathcal{F},\epsilon}(h)$ is the largest subadditive function that minorizes $\min\bracket{J_P(h),\epsilon \Theta_{\mathcal{F}}(h)}$.
\end{lemma}
The consequence of Lemma \ref{penalty-tightness} is that if $\min\bracket{J_P(h),\epsilon \Theta_{\mathcal{F}}(h)}$ is subadditive then $\Lambda_{\mathcal{F},\epsilon}(h) = \min\bracket{J_P(h),\epsilon \Theta_{\mathcal{F}}(h)}$ since a function always minorizes itself. In the proof of Lemma \ref{penalty-tightness}, we show that both $J_P$ and $\epsilon \Theta_{\mathcal{F}}$ are subadditive and so if $\min\bracket{J_P,\epsilon \Theta_{\mathcal{F}}}$ is consistently equal to either $J_P$ or $\epsilon \Theta_{\mathcal{F}}$ for some $\epsilon$ then we have equality.

We now present a necessary and sufficient condition for a function $h:\Omega \to \mathbb{R}$ so that $\Lambda_{\mathcal{F},\epsilon}(h) = \epsilon \Theta_{\mathcal{F}}(h)$ for all $\epsilon > 0$. In doing so, not only do we lead to a better understanding of distributional robustness, we also contribute to understanding tightness of previous results and inequalities subsumed by Corollary \ref{theta-upperbound}. It turns out rather surprisingly that the characterization is directly related to penalty-regularized critic losses.
\begin{theorem}\label{nec-suff-condition}
A function $h \in \mathscr{F}(\Omega, \mathbb{R})$ satisfies $\Lambda_{\mathcal{F}, \epsilon}(h) = \Theta_{\mathcal{F}}(h)$ if and only if
\begin{align}
\label{loss-condition}
    h \in \arginf_{\hat{h} \in \mathscr{F}(\Omega, \mathbb{R})} \bracket{\E_{P}[\hat{h}] - \E_{\mu}[\hat{h}] + \epsilon \Theta_{\mathcal{F}}(\hat{h}) },
\end{align}
for some $\mu \in \mathscr{P}(\Omega)$.
\end{theorem}
First, note that this characterization holds for any $h$ as long as one can find a $\mu$ that satisfies Equation (\ref{loss-condition}). In particular, when $\mu = P$, then the minimizers of Equation (\ref{loss-condition}) are constant functions. Furthermore, Equation (\ref{loss-condition}) can be viewed as a regularized binary classification objective in the following way: $\Omega$ is the input space, $Y = \braces{-1,+1}$ is the label space, $\hat{h}: \Omega \to \mathbb{R}$ is the classifier, $\Theta_{\mathcal{F}}$ is a penalty with weight $\epsilon$, and $P$ (resp. $\mu$) corresponds to the $-1$ (resp. $+1$) class conditional distribution. In particular, this is precisely the objective for the discriminator in penalty-based GANs \citep{gulrajani2017improved,thanh2019improving}, referred to as the critic loss where $P$ is the fake data generated by a model and $\mu$ is the real data. Intuitively, the discriminator function will assign negative values to regions of $\mu$ and positive values to regions of $P$. The discriminator function is then used to guide learning of the model generator by focusing on moving $\mu$ to where $h$ assigns higher values. In conjunction with Theorem \ref{main-ipm-robustness-thm}, this discriminator is robust to shifts to the distribution $P$ and we outline the consequence more clearly in the following Corollary.
\begin{corollary}\label{condition-corr}
Let $P_{+},P_{-} \in \mathscr{P}(\Omega)$ and suppose $\mathcal{F} \subseteq \mathscr{F}(\Omega, \mathbb{R})$ is even. If
\begin{align}\label{loss-condition-2}
    h^{*} \in \arginf_{\hat{h} \in \mathscr{F}(\Omega, \mathbb{R})} \bracket{\E_{P_{-}}[\hat{h}] - \E_{P_{+}}[\hat{h}] + \epsilon \Theta_{\mathcal{F}}(\hat{h}) },
\end{align}
then we have
\begin{align*}
    \inf_{Q \in B_{\epsilon, \mathcal{F}}(P_{+}) } \int_{\Omega} h^{*} dQ = \int_{\Omega} h^{*} dP_{+} - \epsilon \Theta_{\mathcal{F}}(h^{*})\\
    \sup_{Q \in B_{\epsilon, \mathcal{F}}(P_{-}) } \int_{\Omega} h^{*} dQ = \int_{\Omega} h^{*} dP_{-} + \epsilon \Theta_{\mathcal{F}}(h^{*}).
    \end{align*}
\end{corollary}
The implication of this corollary is that the classifier learned by solving Equation (\ref{loss-condition-2}) is still positive (resp. negative) around $B_{\epsilon, \mathcal{F}}$ neighborhoods of $P_{+}$ (resp. $P_{-}$). In the context of GANs, $P_{+}$ and $P_{-}$ will be the real and fake distributions. This is a rather intuitive result since the classifier $h^{*}$ is penalized against $\Theta_{\mathcal{F}}$ however the above Corollary gives formal perspectives along with interpretations to the weighting $\epsilon$ and the choice of penalty (induced by $\mathcal{F}$). We write this Corollary in a more general form since we believe it can be useful for other studies of robustness. An example of this is robustness certification, which cares about distributional shifts to a reference measure for a classifier (see Definition 2.2 of \citep{dvijotham2020framework}). We leave the details of such developments for future work.  
Corollary \ref{condition-corr} uses the fact that the condition outlined in Theorem \ref{nec-suff-condition} is sufficient; however, we emphasize that it is also necessary, suggesting an intimate link between regularized binary-classification and distributional robustness. 
\section{Distributional Robustness of $f$-GANs}
In this section, we show how our main theorem can naturally be applied into the robustness for $f$-GANs more generally. $\Omega$ will typically be a high dimensional Euclidean space to represent the set of images and $P \in \mathscr{P}(\Omega)$ will be an empirical distribution that we are interested in modelling. The model distribution, also referred to as the generative distribution denoted as $\mu \in \mathscr{P}(\Omega)$, is learned by minimizing a divergence between $P$ and $\mu$. We now introduce the $f$-GAN objective, which is a central divergence in the GAN paradigm.
\begin{definition}[$f$-GAN, \citep{nowozin2016f}]
Let $f: \mathbb{R} \to (-\infty, \infty]$ be a lower semicontinuous convex function  with $f(1) = 0$ and $\mathcal{H} \subset \mathscr{F}(\Omega, \operatorname{dom}f^{\star})$ be a set of discriminators. The GAN objective for data $P \in \mathscr{P}(\Omega)$ and model $\mu \in \mathscr{P}(\Omega)$ is
\begin{align*}
    \operatorname{GAN}_{f, \mathcal{H}}(\mu ; P) = \sup_{h \in \mathcal{H}}\bracket{\int_{\Omega}h dP - \int_{\Omega} f^{\star}(h) d\mu }, 
\end{align*}
where $f^{\star}(y) = \sup_{x \in \mathbb{R}} \bracket{x\cdot y - f(x)}$ is the convex conjugate.
\end{definition}
We are interested in minimizing the above objective with respect to $\mu$, which results in a min-max objective due to the supremum taken over $\mathcal{H}$. One should note that there are two components of this objective that characterize it, the function $f$ and discriminator set $\mathcal{H}$. In practice, the discriminator set is often restricted, and so the resulting objective is not a divergence; however, empirical studies have observed convergence \citep{fedus2017many}, which warrants an investigation into the effects of a restricted discriminator on model performance.
Existing work has hinted the benefits of a restricted discriminator, for example, \citep{zhang2017discrimination} show that generalization is related to the Rademacher complexity of the discriminator set and suggest a discrimination-generalization trade-off. Other work has suggested that the particular setting of Lipschitz discriminators leads to improvements for both practical \citep{zhang2017discrimination, fedus2017many,zhou2019lipschitz,wu2019generalization,NIPS2018_7771} and theoretical purposes \citep{husain2019primal, NIPS2018_7771, liu2017approximation}. It is clear that the discriminator set is a key character in the tale of success of GANs; however, the existing literature is silent on the story of robustness, and this is precisely the link we establish.
%ifferent choices of these span a variety of divergences including all the $f$-divergences and Integral Probability Metrics. If one picks $\mathcal{H}$ to be the largest possible, ie. $\mathcal{H} = \mathscr{F}(\Omega, \operatorname{dom}(f^{\star}))$ then $\operatorname{GAN}_{f, \mathcal{H}}(\mu ; P) = d_{f}(P,\mu)$ becomes the $f$-divergence. On the other hand, if one picks $f(x) = \infty \cdot \llbracket x \neq 1 \rrbracket$, which can be considered an extreme choice, then $\operatorname{GAN}_{f, \mathcal{H}}(\mu ; P) = d_{\mathcal{H}}(P,\mu)$ is an IPM. 
Consider now the perspective of distributional robustness, which we formulate with the following objective, letting $\mathcal{G}$ denote the set of generative distributions:
\begin{align*}
  \inf_{\mu \in \mathcal{G}} \sup_{Q \in B_{\epsilon, \mathcal{F}}(P)} \sup_{h \in \mathcal{H}} \bracket{\E_{P}[h] - \E_{\mu}[f^{\star}(h)] }.
\end{align*}
Compared to the standard min-max in GANs between the generator and discriminator, we have a min-max-max where the second max represents an adversary who aids the discriminator in changing the distribution $P$. Notationally, this objective is equivalent to $\inf_{\mu \in \mathcal{G}}\operatorname{GAN}_{f, \mathcal{H}}(\mu ; Q)$. Consider the setting of $f(t) = \infty \cdot \llbracket t \neq 1 \rrbracket$ in which case $\operatorname{GAN}_{f, \mathcal{H}}(\mu ; Q) = d_{\mathcal{H}}(Q,\mu)$ is the IPM. For an extreme choice of $\mathcal{H} = \mathscr{F}(\Omega, \mathbb{R})$, we have $d_{f, \mathcal{H}}(\mu ; Q) = \infty$ if $\mu \neq Q$, which can easily be made large by an adversary. This points to the intuition that a restricted choice for $\mathcal{H}$ is more appropriate for robustness, and this is what we formalize with the following Theorem.
\begin{theorem}
\label{GAN-robust-thm}
Let $f: \mathbb{R} \to \mathbb{R}$ be a convex lower semi-continuous function with $f(1) = 0$, $\mathcal{F} \subseteq \mathscr{F}(\Omega, \mathbb{R})$ and $\mathcal{H} \subseteq \mathscr{F}(\Omega, \operatorname{dom}(f^{\star}))$. For any model and data distributions $\mu, P \in \mathscr{P}(\Omega)$ respectively, we have for all $\epsilon > 0$
\begin{align*}
    \sup_{Q \in B_{\epsilon, \mathcal{F}}(P)} \operatorname{GAN}_{f, \mathcal{H}}(\mu ; Q) \leq \operatorname{GAN}_{f, \mathcal{H}}(\mu ; P) + \epsilon \sup_{h \in \mathcal{H}} \Theta_{\mathcal{F}}(h).
\end{align*}
\end{theorem}
This Theorem tells us that the robust version of the GAN objective can be upper bounded by the standard GAN objective plus a term that quantifies the complexity of the discriminator set. Note that the robustness parameters ($\epsilon$ and $\mathcal{F}$) interact only with the discriminator set and not the generative model $\mu$, revealing the importance of choosing a regularized discriminator set $\mathcal{H}$. To see this more clearly, consider the setting $\mathcal{F} = \mathcal{H}$, and since $\Theta_{\mathcal{H}}(h) \leq 1$, we have
\begin{align}\label{gan-imp-eq}
    \sup_{Q \in B_{\epsilon, \mathcal{H}}(P)} \operatorname{GAN}_{f, \mathcal{H}}(\mu ; Q) \leq \operatorname{GAN}_{f, \mathcal{H}}(\mu ; P) + \epsilon,
\end{align}
for all $\epsilon > 0$. The key insight is that training GANs using discriminators $\mathcal{H}$ yields guarantees on the robust GAN objective for adversaries who pick $Q$ from $B_{\epsilon, \mathcal{H}}(P)$. From the previous discussion, if one picks discriminators $\mathcal{H}$ that are too strong then the ball $B_{\epsilon, \mathcal{H}}(P)$ will shrink and become singleton $\braces{P}$ when $\mathcal{H} = \mathscr{F}(\Omega, \mathbb{R})$. On the other hand, if $\mathcal{H}$ is chosen to be smaller then the uncertainty set is larger; however, the first term $\operatorname{GAN}_{f, \mathcal{H}}$ will be a weaker divergence, since the discriminator set determines the strength of the objective \citep{liu2017approximation}. Hence, there is a trade-off between discrimination and robustness, that complements and parallels the discrimination-generalization story described in \citep{zhang2017discrimination}.

We now discuss the particular settings of $\mathcal{F}$ and how our theorem gives a perspective of distributional robustness on existing GAN methods. First, consider choices of $\mathcal{F}$ so that $d_{\mathcal{F}}$ corresponds to MMD, Fisher IPM and Sobelov IPM which translates to the MMD-GAN, Fisher-GAN and Sobelov GAN respectively, allowing us to view these methods from a robustness perspective in light of Theorem \ref{GAN-robust-thm} and Equation (\ref{gan-imp-eq}). Furthermore, our result also contributes to the positive commentary under the popular choice of Lipschitz regularized discriminators, guarantees against adversaries selecting from Wasserstein uncertainty sets.
It should be noted that recently, a method that regularizes discriminators by minimizing a penalty referred to as $0$-GP \citep{thanh2019improving} has proven convergence and generalization guarantees. It can be easily shown that this penalty satisfies the conditions of Lemma \ref{hom-penalties} for $k = 2$ due to its resemblance to the Sobelov IPM, allowing us to present a robustness interpretation for this penalty. 
\section{Conclusion}
Our results extend the Distributionally Robust Optimization (DRO) framework to IPMs, which reveal further importance of the role regularization plays for robustness and machine learning at large. Unlike most DRO applications to machine learning, we present equality and show that achieving this is fundamentally rooted in regularized binary classification. We then show that DRO can be extended to understand GANs and unveil the role of discrimination regularization in these frameworks. The results will also help DRO explain regularization penalties through the lens of robustness in the future. Our contributions are modular and pave the way to build on related areas, one such example being robustness certification, which we leave for the subject of future work.
\section*{Acknowledgements}
We would like to thank Jeremias Knoblauch for his helpful suggestions on improving clarity and presentation.

\bibliographystyle{apalike}
\bibliography{references}

\clearpage
\newpage
%\onecolumn
\section*{Proofs of Main Results}\label{supp-formal}
Before we begin, we introduce some notation that will be used to prove the main results that is exclusive to the Appendix. We will be invoking general convex analysis on the space $\mathscr{F}(\Omega, \mathbb{R})$, in the same fashion as \citep{liu2018inductive}, noting that $\mathscr{F}(\Omega, \mathbb{R})$ is a Hausdorff locally convex space (through the uniform norm). We use $\mathscr{B}(\Omega)$ to denote the denote the set of all bounded and finitely additive signed measures over $\Omega$ (with a given $\sigma$-algebra). For any set $D \subseteq \mathscr{B}(\Omega)$ and $h \in \mathscr{F}(\Omega, \mathbb{R})$, we use $\sigma_{D}(h) = \sup_{\nu \in D} \ip{h}{\nu}$ and $\delta_{D}(\nu) = \infty \cdot \llbracket \nu \notin D \rrbracket$ to denote the \textit{support} and \textit{indicator} functions such as in \citep{rockafellar1970convex}. We introduce the conjugate specific to these spaces
\begin{definition}[\citep{rockafellar1968integrals}]
For any proper convex function $F: \mathscr{F}(\Omega, \mathbb{R}) \to (-\infty, \infty)$, we have for any $\mu \in \mathscr{B}(\Omega)$ we define
\begin{align*}
    F^{\star}(\mu) = \sup_{h \in \mathscr{F}} \bracket{ \int_{\Omega} h d\mu - F(h) }
\end{align*}
and for any $h \in \mathscr{F}(\Omega, \mathbb{R})$ we define
\begin{align*}
    F^{\star \star}(h) = \sup_{\mu \in \mathscr{B}(\Omega)} \bracket{ \int_{\Omega} h d\mu - F^{\star}(\mu) }.
\end{align*}
\end{definition}
\begin{theorem}[\citep{zalinescu2002convex} Theorem 2.3.3]
\label{self-conjugacy}
If $X$ is a Hausdorff locally convex space, and $F: X \to (-\infty, \infty]$ is a proper lower semi-continuous function then $F^{\star \star} = F$.
\end{theorem}
There is an additional robustness result which we will deploying for several proofs which holds for any space $A$ that admits Polish topology.
\begin{lemma}
\label{ipm-closed-hull}
For any $\mathcal{F} \subseteq \mathscr{F}(\Omega, \mathbb{R})$, we have that
\begin{align*}
     d_{\mathcal{F}}(P,\mu) = d_{\chull{\mathcal{F}}}(P,\mu).
\end{align*}
\end{lemma}
  \begin{proof}
  Let $\Delta_n := \braces{ \alpha \in [0,1]^n : \sum_{i=1}^n \alpha = 1 }$ Note that we have
  \begin{align*}
      d_{\operatorname{co}\bracket{\mathcal{F}}}(P,\mu) &= \sup_{n \in \mathbb{N}, \alpha \in \Delta_n, f_i \in \mathcal{F} \forall i=1,\ldots,n} \braces{\E_{P}\left[\sum_{i=1}^n \alpha_i f_i \right] - \E_{\mu}\left[\sum_{i=1}^n \alpha_i f_i \right]}\\
      &= \sup_{n \in \mathbb{N}, \alpha \in \Delta_n, f_i \in \mathcal{F} \forall i=1,\ldots,n}\sum_{i=1}^n \alpha_i \braces{\E_{P}\left[f_i \right] - \E_{\mu}\left[ f_i \right]}\\
      &= \sup_{n \in \mathbb{N}, \alpha \in \Delta_n} \sum_{i=1}^n \alpha_i \sup_{f_i \in \mathcal{F}} \braces{\E_{P}\left[f_i \right] - \E_{\mu}\left[ f_i \right]}\\
      &= \sup_{n \in \mathbb{N}, \alpha \in \Delta_n} \sum_{i=1}^n \alpha_i d_{\mathcal{F}}(P,\mu)\\
      &= d_{\mathcal{F}}(P,\mu)
  \end{align*}
  It is also closed under taking the closure since $d_{\mathcal{F}}$ is the supremum of continuous (linear) functions and the supremum over a set with a linear objective is equal to taking the supremum over the closure of that set. 
  \end{proof}

\begin{definition}
For any $\mathcal{F} \subseteq \mathscr{F}(\Omega,\mathbb{R})$, we define the functional $R_{\mathcal{F}} : \mathscr{F}(\Omega, \mathbb{R}) \to [0,\infty]$ as
\begin{align*}
    R_{\mathcal{F}}(h) := \int_{\Omega} h dP + \infty \cdot \llbracket h \notin \chull{\mathcal{F}} \rrbracket.
\end{align*}
\end{definition}
\begin{lemma}
\label{ipm-proper}
For any $\mathcal{F} \subseteq \mathscr{F}(\Omega, \mathbb{R})$, $\mathcal{R}_{\mathcal{F}}$ is proper convex and lower semi-continuous.
\end{lemma}
\begin{proof}
The mapping $h \mapsto \int_{\Omega} h dP$ is clearly convex and lower semi-continuous. Since $\chull{\mathcal{F}}$ is a closed and convex set, the indicator function $\infty \cdot \llbracket h \notin \chull{\mathcal{F}} \rrbracket$ is proper convex and lower semi-continuous and thus the result follows.
\end{proof}
\begin{lemma}
\label{conjugate-of-ipm}
The mappings $\nu \mapsto d_{\mathcal{F}}(\nu,P)$ and $h \mapsto R_{\mathcal{F}}(h)$ are convex conjugates 
\end{lemma}
\begin{proof}
Note first that for any $\nu \in \mathscr{B}(\Omega)$
\begin{align*}
    R_{\mathcal{F}}^{\star}(\nu) &= \sup_{h \in \mathscr{F}(\Omega, \mathbb{R})} \braces{ \int_{\Omega} h d\nu - \int_{\Omega} h dP - \infty \cdot \llbracket h \notin \chull{\mathcal{F}}\rrbracket }\\
    &= \sup_{h \in \chull{\mathcal{F}}} \braces{ \int_{\Omega} h d\nu - \int_{\Omega} h dP}\\
    &= d_{ \chull{\mathcal{F}}}(\nu, P)\\
    &\stackrel{(1)}{=} d_{ \mathcal{F}}(\nu,P),
\end{align*}
where $(1)$ is due to Lemma \ref{ipm-closed-hull}. We also have that
\begin{align*}
    \bracket{d_{\mathcal{F}}(\cdot, P)}^{\star}(h) &= \sup_{\nu \in \mathscr{B}(\Omega)} \braces{ \int_{\Omega} h d\nu - d_{\mathcal{F}}(\nu,P) }\\
    &\stackrel{(1)}{=} \sup_{\nu \in \mathscr{B}(\Omega)} \braces{ \int_{\Omega} h d\nu - R_{\mathcal{F}}^{\star}(\nu) }\\ 
    &\stackrel{(2)}{=} R_{\mathcal{F}}^{\star \star}(\nu)\\
    &\stackrel{(3)}{=} R_{\mathcal{F}}(\nu),
\end{align*}
where $(1)$ holds due to the above, $(2)$ holds by definition of conjugate and $(3)$ holds by a combination of Lemma \ref{self-conjugacy} and Lemma \ref{ipm-proper}.
\end{proof}
We also present a lemma which will prove to be useful in proving the main results.
\begin{lemma}
\label{theta-convex}
For any $\mathcal{F} \subset \mathscr{F}(\Omega, \mathbb{R})$, the mapping $h \mapsto \Theta_{\mathcal{F}}(h)$ is convex.
\end{lemma}
\begin{proof}
First notice that for any $t > 0$ and $h \in \mathscr{F}(\Omega, \mathbb{R})$ we have that $\Theta_{\mathcal{F}}(t \cdot h) = t \cdot \Theta_{\mathcal{F}}(h)$. For any $t \in [0,1]$ and $h,h' \in \mathscr{F}(\Omega, \mathbb{R})$, consider the element $\tilde{h} := t \cdot h + (1-t) \cdot h'$. Since $t \cdot h \in t \Theta_{\mathcal{F}}(h) \cdot \chull{\mathcal{F}}$ and $(1-t)h \in (1-t) \Theta_{\mathcal{F}}(h) \cdot \chull{\mathcal{F}}$, we have that
\begin{align*}
    \tilde{h} \in t \Theta_{\mathcal{F}}(h) \cdot \chull{\mathcal{F}} + (1-t) \Theta_{\mathcal{F}}(h) \cdot \chull{\mathcal{F}}\\
    \iff \tilde{h} \in \bracket{t \Theta_{\mathcal{F}}(h)  + (1-t) \Theta_{\mathcal{F}}(h')} \cdot \chull{\mathcal{F}},
\end{align*}
which in turn implies that $\Theta_{\mathcal{F}}(\tilde{h}) \leq t \Theta_{\mathcal{F}}(h)  + (1-t) \Theta_{\mathcal{F}}(h')$, proving convexity of $\Theta_{\mathcal{F}}$.
\end{proof}
\subsection{Proof of Theorem \ref{main-ipm-robustness-thm}} \label{proof:main-ipm-robustness-thm}
\begin{theorem}
Let $\mathcal{F} \subseteq \mathscr{F}(\Omega, \mathbb{R})$ and $P \in \mathscr{P}(\Omega)$. For any $h \in \mathscr{F}(\Omega, \mathbb{R})$ and for all $\epsilon > 0$
\begin{align*}
    \sup_{Q \in B_{\epsilon, \mathcal{F}}(P)} \int_{\Omega} h dQ = \int_{\Omega} h dP + \Lambda_{\mathcal{F}, \epsilon}(h).
\end{align*}
\end{theorem}
\begin{proof}
We first require two lemmata.
\begin{lemma}\label{IPM-Lemma-1}
For any $\mathcal{F} \subseteq \mathscr{F}(\Omega, \mathbb{R})$, $P \in \mathscr{P}(\Omega)$, $\lambda \geq 0$ and $h \in \mathscr{F}(\Omega, \mathbb{R})$, we have
\begin{align*}
    \sup_{Q \in \mathscr{P}(\Omega)} \bracket{\int_{\Omega} h dQ - \lambda d_{\mathcal{F}}(P,Q) } = R_{\lambda \mathcal{F}} \iconv \sigma_{\mathscr{P}(\Omega)}(h)
\end{align*}
\end{lemma}
\begin{proof}
We use a standard result from convex analysis which states that the convex conjugate of the sum of two functions is the infimal convolution of their conjugates. Hence we have
\begin{align*}
    \sup_{Q \in \mathscr{P}(\Omega)} \bracket{\int_{\Omega} h dQ - \lambda d_{\mathcal{F}}(P,Q) } &= \sup_{Q \in \mathscr{B}(\Omega)} \bracket{\int_{\Omega} h dQ - \lambda d_{\mathcal{F}}(P,Q) - \infty \cdot \llbracket Q \notin \mathscr{P}(\Omega) \rrbracket}\\
    &=  \bracket{\lambda d_{\mathcal{F}}(P,Q) + \infty \cdot \llbracket Q \notin \mathscr{P}(\Omega) \rrbracket}^{\star}\\
    &= \bracket{\lambda d_{\mathcal{F}}(P,Q}^{\star} \iconv \bracket{\infty \cdot\llbracket Q \notin \mathscr{P}(\Omega) \rrbracket}^{\star}\\
    &= R_{\lambda \mathcal{F}} \iconv \sigma_{\mathscr{P}(\Omega)}(h),
\end{align*}
which follows from Lemma \ref{conjugate-of-ipm} and the fact that support functions are conjugates of indicator functions \cite[Section 3.4.1, Example (a)]{penot2012calculus}.
\end{proof}
\begin{lemma} \label{IPM-Lemma-2}
For any $\mathcal{F} \subseteq \mathscr{F}(\Omega, \mathbb{R})$, $P \in \mathscr{P}(\Omega)$, and $h \in \mathscr{F}(\Omega, \mathbb{R})$, we have
\begin{align*}
    \inf_{\lambda \geq 0}\bracket{ R_{\lambda \mathcal{F}} \iconv \sigma_{\mathscr{P}(\Omega)}(h)+\lambda\epsilon} = \int_{\Omega} h dP +  J_{P} \iconv \epsilon  \Theta_{\mathcal{F}}(h)
\end{align*}
\end{lemma}
\begin{proof}
Using the definition of infimal convolution, we have
\begin{align*}
    &\inf_{\lambda \geq 0}\bracket{ R_{\lambda\mathcal{F}} \iconv \sigma_{\mathscr{P}(\Omega)}(h)+\lambda\epsilon}\\ &= \inf_{\lambda \geq 0}\bracket{ \inf_{h' \in \mathscr{F}(\Omega, \mathbb{R})} \bracket{\int_{\Omega} (h - h') dP + \infty \cdot \llbracket h - h' \notin \chull{\lambda \mathcal{F}} \rrbracket +\sigma_{\mathscr{P}(\Omega)} (h) } +\lambda\epsilon}\\
    &= \inf_{\lambda \geq 0} \inf_{h' \in \mathscr{F}(\Omega, \mathbb{R})} \bracket{ \int_{\Omega}h dP -\int_{\Omega} h' dP + \infty \cdot \llbracket h - h' \notin \chull{\lambda \mathcal{F}} \rrbracket +\sigma_{\mathscr{P}(\Omega)} (h')  + \lambda \epsilon}\\
    &= \int_{\Omega} h dP + \inf_{h' \in \mathscr{F}(\Omega, \mathbb{R})}\bracket{ - \int_{\Omega}h' dP +  \inf_{\lambda \geq 0}\bracket{\infty \cdot \llbracket h - h' \notin \chull{\lambda \mathcal{F}} \rrbracket  + \lambda \epsilon }  +\sigma_{\mathscr{P}(\Omega)} (h') }\\
    &= \int_{\Omega} h dP + \inf_{h' \in \mathscr{F}(\Omega, \mathbb{R})}\bracket{\sigma_{\mathscr{P}(\Omega)} (h') - \int_{\Omega}h' dP +  \inf_{\lambda \geq 0}\bracket{\infty \cdot \llbracket h - h' \notin \chull{\lambda \mathcal{F}} \rrbracket  + \lambda \epsilon }  }\\
    &= \int_{\Omega} h dP + \inf_{h' \in \mathscr{F}(\Omega, \mathbb{R})}\bracket{\sigma_{\mathscr{P}(\Omega)} (h') - \int_{\Omega}h' dP +  \inf_{\lambda \geq 0}\bracket{\infty \cdot \llbracket h - h' \notin \lambda \cdot \chull{ \mathcal{F}} \rrbracket  + \lambda \epsilon }  }\\
    &= \int_{\Omega} h dP + \inf_{h' \in \mathscr{F}(\Omega, \mathbb{R})}\bracket{J_{P}(h') +  \epsilon \Theta_{\mathcal{F}}(h-h')  }\\
    &= \int_{\Omega} h dP + J_{P} \iconv \epsilon  \Theta_{\mathcal{F}}(h).
\end{align*}
\end{proof}
We are now ready to prove the Theorem. By introducing a dual variable $\lambda > 0$ that penalizes the ball constraint, we have 
\begin{align*}
    \sup_{Q \in B_{\epsilon, \mathcal{F}}(P)} \int_{\Omega} hdQ &= \sup_{Q \in \mathscr{P}(\Omega) : d_{\mathcal{F}}(Q,P) \leq \epsilon} \int_{\Omega} hdQ\\
    &= \sup_{Q \in \mathscr{P}(\Omega)} \inf_{\lambda \geq 0} \bracket{ \int_{\Omega} h dQ + \lambda \bracket{\epsilon - d_{\mathcal{F}}(Q,P) }}\\
    &\stackrel{(1)}{=}  \inf_{\lambda \geq 0} \sup_{Q \in \mathscr{P}(\Omega)} \bracket{ \int_{\Omega} h dQ + \lambda \bracket{\epsilon - d_{\mathcal{F}}(Q,P) }}\\
    &= \inf_{\lambda \geq 0} \bracket{\sup_{Q \in \mathscr{P}(\Omega)} \bracket{\int_{\Omega}h dQ - \lambda d_{\mathcal{F}}(Q,P)} + \lambda \epsilon}\\
    &\stackrel{(2)}{=} \inf_{\lambda \geq 0} \bracket{R_{\lambda \mathcal{F}} \iconv \sigma_{\mathscr{P}(\Omega)}(h) + \lambda \epsilon}\\
    &\stackrel{(3)}{=} \int_{\Omega} hdP + J_P \iconv \epsilon \Theta_{\mathcal{F}}(h),
\end{align*}
where $(2)$ and $(3)$ hold due to Lemma \ref{IPM-Lemma-1} and \ref{IPM-Lemma-2} respectively. To see why $(1)$ holds, first note that the mapping $Q \mapsto \int_{\Omega} h dQ + \lambda \bracket{\epsilon - d_{\mathcal{F}}(Q,P) }$ is concave and lower semicontinuous since $d_{\mathcal{F}}$ is the supremum of linear functions. Next we have by an application of the Banach-Alaogu Theorem that $\mathscr{P}(\Omega)$ is compact \cite[Lemma 27 (b)]{liu2018inductive}. Hence by \cite[Theorem 2]{fan1953minimax}, (1) follows.
\end{proof}
\subsection{Proof of Corollary \ref{theta-upperbound}} \label{proof:theta-upperbound}
\begin{corollary}
Let $\mathcal{F} \subseteq \mathscr{F}(\Omega, \mathbb{R})$ and $P \in \mathscr{P}(\Omega)$. For any $h \in \mathscr{F}(\Omega, \mathbb{R})$ and for all $\epsilon > 0$
\begin{align*}
\sup_{Q \in B_{\epsilon, \mathcal{F}}(P)} \int_{\Omega} h dQ \leq \int_{\Omega} h dP + \epsilon \inf_{b \in \mathbb{R}}\Theta_{\mathcal{F}}(h - b).
\end{align*}
\end{corollary}
\begin{proof}
By definition of the infimal convolution we can consider a decomposition of the form $h_1 = b$ and $h_2 = h - b$ for some $b \in \mathbb{R}$. notice that $J_P(b) = 0$ and by taking the smallest possible $b \in \mathbb{R}$ yields
\begin{align*}
    \Theta_{\mathcal{F}, \epsilon}(h) \leq \epsilon \inf_{b \in \mathbb{R}} \Theta_{\mathcal{F}}(h - b),
\end{align*}
which completes the proof.
\end{proof}
\subsection{Proof of Lemma \ref{hom-penalties}}\label{proof:hom-penalties}
\begin{lemma}
Let $\zeta: \mathscr{F}(\Omega, \mathbb{R}) \to [0,\infty]$ be a penalty such that $\zeta(a \cdot h) = a^k \cdot \zeta(h)$ for any $h \in \mathscr{F}(\Omega, \mathbb{R})$, $k,a > 0$. Let $\mathcal{F} = \braces{h : \zeta(h) \leq 1}$ then we have $\Theta_{\mathcal{F}}(h) \leq \sqrt[k]{\zeta(h)}$ with equality if $\zeta$ is convex.
\end{lemma}
\begin{proof}
Let us consider the non-convex case so that $\mathcal{F}$ is not necessarily convex. We then have for any $\mathcal{F} \subseteq \mathscr{F}(\Omega, \mathbb{R})$
\begin{align*}
    h \in \chull{\lambda \mathcal{F}} &\iff h \in \lambda \chull{ \mathcal{F}}\\
                                      &\iff \frac{h}{\lambda} \in \chull{\mathcal{F}}
\end{align*}
For a fixed $h \in \mathscr{F}(\Omega, \mathbb{R})$, set $\lambda = \sqrt[k]{\zeta(h)}$ and notice that 
\begin{align*}
    \zeta\bracket{\frac{h}{\lambda}} &= \zeta\bracket{\frac{h}{\sqrt[k]{\zeta(h)}}}\\
                                        &= \bracket{\frac{1}{\sqrt[k]{\zeta(h)}}}^k \zeta\bracket{h} \\
                                        &= \zeta\bracket{h},
\end{align*}
and so we have $\Theta_{\mathcal{F}}(h) \leq \sqrt[k]{\zeta(h)}$. In the case when the penalty is convex, we have that $\mathcal{F}$ will be convex and so
\begin{align*}
         h \in \lambda \chull{ \mathcal{F}} &\iff \frac{h}{\lambda} \in \chull{\mathcal{F}}\\
                                      &\iff \frac{h}{\lambda} \in \mathcal{F}\\
                                      &\iff \zeta\bracket{\frac{h}{\lambda}} \leq 1\\
                                      &\iff \frac{1}{\lambda^k} \zeta(h) \leq 1\\
                                      &\iff \zeta(h) \leq \lambda^k\\
                                      &\iff \sqrt[k]{\zeta(h)} \leq \lambda. 
\end{align*}
Hence we have $\Theta_{\mathcal{F}}(h) = \inf_{\sqrt[k]{\zeta(h)} \leq \lambda} \lambda = \sqrt[k]{\zeta(h)}$.
\end{proof}
\subsection{Proof of Lemma \ref{penalty-tightness}}\label{proof:penalty-tightness}
\begin{lemma}
The mapping $h \mapsto \Lambda_{\mathcal{F},\epsilon}(h)$ is subadditive and $\Lambda_{\mathcal{F},\epsilon}(h)$ is the largest subadditive function that minorizes $\min\bracket{J_P(h),\epsilon \Theta_{\mathcal{F}}(h)}$.
\end{lemma}
\begin{proof}
Since $\Theta_{\mathcal{F}}(h)$ is convex (Lemma \ref{theta-convex}) and $\Theta_{\mathcal{F}}(t \cdot h) = t \cdot \Theta_{\mathcal{F}}(h)$  for $t > 0$, it follows that $\Theta_{\mathcal{F}}(h)$ is subadditive. Next notice that $J_P$ is subadditive since for any $h, h' \in \mathscr{F}(\Omega, \mathbb{R})$
\begin{align*}
    J_P(h + h') &= \sup_{\omega \in \Omega} h(\omega) + h'(\omega) - \int_{\Omega} h dP - \int_{\Omega} h' dP\\
    &\leq  \sup_{\omega \in \Omega} h(\omega) - \int_{\Omega} h dP + \sup_{\omega \in \Omega} h'(\omega) - \int_{\Omega} h' dP\\
    &= J_P(h) + J_P(h').
\end{align*}
Next notice that $J_P(0) = 0$ and $\epsilon \Theta_{\mathcal{F}}(0) = 0$. By \cite[Theorem~2.5(c)]{stromberg1994study} we have that $\Lambda_{\mathcal{F}, \epsilon}$ is sub-additive and that it is the largest subadditive function that minorizes $\min\bracket{J_P(h), \epsilon \Theta_{\mathcal{F}}(h)}$.
\end{proof}
\subsection{Proof of Theorem \ref{nec-suff-condition}}\label{proof:nec-suff-condition}
\begin{theorem}
A function $h \in \mathscr{F}(\Omega, \mathbb{R})$ satisfies $\Lambda_{\mathcal{F}, \epsilon}(h) = \Theta_{\mathcal{F}}(h)$ if and only if
\begin{align*}
    h \in \arginf_{\hat{h} \in \mathscr{F}(\Omega, \mathbb{R})} \bracket{\E_{P}[\hat{h}] - \E_{\mu}[\hat{h}] + \epsilon \Theta_{\mathcal{F}}(\hat{h}) },
\end{align*}
for some $\mu \in \mathscr{P}(\Omega)$.
\end{theorem}
\begin{proof}
To prove this Theorem, we use the conditions for an optimal decomposition of an infimal convolution as shown in \cite[Lemma~1]{niyobuhungiro2013optimal}. First note that $J_P$ and $\Theta_{\mathcal{F}}$ are convex (Lemma \ref{theta-convex}). Note that the property is equivalent to showing that the decomposition $h_1 = 0$ and $h_2 = h$ is optimal. By \cite[Lemma~1]{niyobuhungiro2013optimal}, this decomposition is optimal if and only if there exists a measure $\nu^{*} \in \mathscr{B}(\Omega)$ such that
\begin{align}
    J_P(0) &= \ip{\nu^{*}}{0} - J_{P}^{\star}(\nu^{*})\label{condition:jp}\\
    \epsilon \Theta_{\mathcal{F}} (h) &= \ip{\nu^{*}}{h} - (\epsilon \Theta_{\mathcal{F}})^{\star}(\nu^{*})\label{condition:theta}
\end{align}
First note that $J_P(h) = \sigma_{\mathscr{P}(\Omega)}(h) + \sigma_{\braces{-P}}(h)$ and using properties of infimal convolutions, we have for any $\nu \in \mathscr{P}(\Omega)$
\begin{align*}
    J_{P}^{\star}(\nu) &= \bracket{\sigma_{\mathscr{P}(\Omega)} + \sigma_{\braces{-P}}}^{\star}(\nu)\\
                       &= \bracket{\sigma_{\mathscr{P}(\Omega)}^{\star} \iconv \sigma_{\braces{-P}}^{\star}}(\nu)\\
                       &= \bracket{\delta_{\mathscr{P}(\Omega)} \iconv \delta_{\braces{-P}}}(\nu)\\
                       &= \inf_{\nu' \in \mathscr{B}(\Omega)} \bracket{\delta_{\mathscr{P}}(\nu') + \delta_{ \braces{-P}}(\nu - \nu') }\\
                       &= \inf_{\nu' \in \mathscr{P}(\Omega)} \delta_{ \braces{-P}}(\nu - \nu') \\
                       &= \infty \cdot \llbracket P + \nu \notin \mathscr{P}(\Omega) \rrbracket\\
                       &= \infty \cdot \llbracket \nu \notin \mathscr{P}(\Omega) - P\rrbracket.
\end{align*}
Since $J_P(0) = \ip{\nu}{0} = 0$ for any $\nu \in \mathscr{B}(\Omega)$, this tells us that a $\nu^{*}$ satisfies the condition of Equation \ref{condition:jp} if and only if $\nu^{*}$ is of the form $\mu - P$ where $\mu$ is any element of $\mathscr{P}(\Omega)$. We can re-arrange Equation \ref{condition:theta} into
\begin{align*}
\ip{\nu^{*}}{h} - \epsilon \Theta_{\mathcal{F}}(h) = (\epsilon \Theta_{\mathcal{F}})^{\star}(\nu^{*}),
\end{align*}
and by definition since $(\epsilon \Theta_{\mathcal{F}})^{\star}(\nu^{*}) = \sup_{\hat{h} \in \mathscr{F}(\Omega, \mathbb{R})}\bracket{ \ip{\nu^{*}}{\hat{h}} - \epsilon \Theta_{\mathcal{F}}(\hat{h}) }$, Equation \ref{condition:theta} setting $\nu^{*} = \mu - P$ becomes
\begin{align}
    &\ip{\nu^{*}}{h} - \epsilon \Theta_{\mathcal{F}}(h) = \sup_{\hat{h} \in \mathscr{F}(\Omega, \mathbb{R})}\bracket{ \ip{\nu^{*}}{\hat{h}} - \epsilon \Theta_{\mathcal{F}}(\hat{h}) }\\ &\iff \ip{\mu - P}{h} - \epsilon \Theta_{\mathcal{F}}(h) = \sup_{\hat{h} \in \mathscr{F}(\Omega, \mathbb{R})}\bracket{ \ip{\mu - P}{\hat{h}} - \epsilon \Theta_{\mathcal{F}}(\hat{h}) }\nonumber \\
    &\iff \E_{\mu}[h] - \E_{P}[h] - \epsilon \Theta_{\mathcal{F}}(h) = \sup_{\hat{h} \in \mathscr{F}(\Omega, \mathbb{R})}\bracket{ \E_{\mu}[\hat{h}] - \E_{P}[\hat{h}] - \epsilon \Theta_{\mathcal{F}}(\hat{h}) }\nonumber\\
    &\iff h \in \argsup_{\hat{h} \in \mathscr{F}(\Omega, \mathbb{R})}\bracket{ \E_{\mu}[\hat{h}] - \E_{P}[\hat{h}] - \epsilon \Theta_{\mathcal{F}}(\hat{h}) }\nonumber\\
    &\iff h \in \arginf_{\hat{h} \in \mathscr{F}(\Omega, \mathbb{R})} \bracket{\E_{P}[\hat{h}] - \E_{\mu}[\hat{h}] + \epsilon \Theta_{\mathcal{F}}(\hat{h}) }. \label{proof:nec-suf-cond}
\end{align}
Hence the decomposition $h_1 = 0$ and $h_2 = h$ is optimal if and only if $h$ satisfies Equation \ref{proof:nec-suf-cond} for some $\mu \in \mathscr{P}(\Omega)$, which is precisely the statement of the Theorem.
\end{proof}
\subsection{Proof of Corollary \ref{condition-corr}}\label{proof:condition-corr}
\begin{corollary}
Let $P_{+},P_{-} \in \mathscr{P}(\Omega)$ and suppose $\mathcal{F} \subseteq \mathscr{F}(\Omega, \mathbb{R})$ is even. If
\begin{align*}
    h^{*} \in \arginf_{\hat{h} \in \mathscr{F}(\Omega, \mathbb{R})} \bracket{\E_{P_{-}}[\hat{h}] - \E_{P_{+}}[\hat{h}] + \epsilon \Theta_{\mathcal{F}}(\hat{h}) },
\end{align*}
then we have
\begin{align*}
    \inf_{Q \in B_{\epsilon, \mathcal{F}}(P_{+}) } \int_{\Omega} h^{*} dQ = \int_{\Omega} h^{*} dP_{+} - \epsilon \Theta_{\mathcal{F}}(h^{*})\\
    \sup_{Q \in B_{\epsilon, \mathcal{F}}(P_{-}) } \int_{\Omega} h^{*} dQ = \int_{\Omega} h^{*} dP_{-} + \epsilon \Theta_{\mathcal{F}}(h^{*})
    \end{align*}
\end{corollary}
\begin{proof}
Applying Theorem \ref{nec-suff-condition} with $P = P_{-}$ and $\mu = P_{+}$ and using Theorem \ref{main-ipm-robustness-thm} yields the result on $B_{\epsilon, \mathcal{F}}(P_{-})$. Notice that $\mathcal{F}$ is even, which means that $\Theta_{\mathcal{F}}(h) = \Theta_{\mathcal{F}}(-h)$ and so we have
\begin{align*}
    &h^{*} \in \arginf_{\hat{h} \in \mathscr{F}(\Omega, \mathbb{R})} \bracket{\E_{P_{-}}[\hat{h}] - \E_{P_{+}}[\hat{h}] + \epsilon \Theta_{\mathcal{F}}(\hat{h}) }\\
    &\iff -h^{*} \in \arginf_{-\hat{h} \in \mathscr{F}(\Omega, \mathbb{R})} \bracket{-\E_{P_{-}}[\hat{h}] + \E_{P_{+}}[\hat{h}] + \epsilon \Theta_{\mathcal{F}}(-\hat{h}) }\\
    &\iff -h^{*} \in \arginf_{-\hat{h} \in \mathscr{F}(\Omega, \mathbb{R})} \bracket{  \E_{P_{+}}[\hat{h}] -\E_{P_{-}}[\hat{h}] + \epsilon \Theta_{\mathcal{F}}(\hat{h}) }.
\end{align*}
We can then apply Theorem \ref{nec-suff-condition} to $-h^{*}$ which means $\Lambda_{\epsilon, \mathcal{F}}(-h^{*}) = \epsilon\Theta_{\mathcal{F}}(-h^{*}) = \epsilon\Theta_{\mathcal{F}}(h^{*})$. Putting this together and applying Theorem \ref{main-ipm-robustness-thm} to $-h^{*}$ gives
\begin{align*}
    \sup_{Q \in B_{\epsilon, \mathcal{F}}(P_{+})} \int_{\Omega} -h^{*} dQ = \int_{\Omega} -h^{*} dP_{+} + \epsilon \Theta_{\mathcal{F}}(h^{*}),
\end{align*}
and multiplying both sides by $-1$ concludes the proof.
\end{proof}
\subsection{Proof of Theorem \ref{GAN-robust-thm}}\label{proof:GAN-robust-thm}
\begin{theorem}
Let $f: \mathbb{R} \to \mathbb{R}$ be a convex lower semi-continuous function with $f(1) = 0$, $\mathcal{F} \subseteq \mathscr{F}(\Omega, \mathbb{R})$ and $\mathcal{H} \subseteq \mathscr{F}(\Omega, \operatorname{dom}(f^{\star}))$. For any model and data distributions $\mu, P \in \mathscr{P}(\Omega)$ respectively, we have for all $\epsilon > 0$
\begin{align*}
    \sup_{Q \in B_{\epsilon, \mathcal{F}}(P)} \operatorname{GAN}_{f, \mathcal{H}}(\mu ; Q) \leq \operatorname{GAN}_{f, \mathcal{H}}(\mu ; P) + \epsilon \sup_{h \in \mathcal{H}} \Theta_{\mathcal{F}}(h)
\end{align*}
\end{theorem}
\begin{proof}
We have
\begin{align*}
    \sup_{Q \in B_{\epsilon, \mathcal{F}}(P)} \operatorname{GAN}_{f, \mathcal{H}}(\mu ; Q) &= \sup_{Q \in B_{\epsilon, \mathcal{F}}(P)}  \sup_{h \in \mathcal{H}}\bracket{\int_{\Omega}h dQ - \int_{\Omega} f^{\star}(h) d\mu }\\
    &\stackrel{(1)}{=}   \sup_{h \in \mathcal{H}}\sup_{Q \in B_{\epsilon, \mathcal{F}}(P)} \bracket{\int_{\Omega}h dQ - \int_{\Omega} f^{\star}(h) d\mu }\\
    &= \sup_{h \in \mathcal{H}} \bracket{ \sup_{Q \in B_{\epsilon, \mathcal{F}}(P)}\int_{\Omega}h dQ - \int_{\Omega} f^{\star}(h) d\mu }\\
    &\stackrel{(2)}{=} \sup_{h \in \mathcal{H}} \bracket{ \int_{\Omega} h dP + \Lambda_{\mathcal{F}, \epsilon}(h) - \int_{\Omega} f^{\star}(h) d\mu }\\
    &\stackrel{(3)}{\leq} \sup_{h \in \mathcal{H}} \bracket{ \int_{\Omega} h dP + \epsilon \Theta_{\mathcal{F}}(h) - \int_{\Omega} f^{\star}(h) d\mu }\\
    &\stackrel{(4)}{\leq} \sup_{h \in \mathcal{H}} \bracket{ \int_{\Omega} h dP  - \int_{\Omega} f^{\star}(h) d\mu } + \epsilon \sup_{h \in \mathcal{H}} \Theta_{\mathcal{F}}(h)\\
    &= \operatorname{GAN}_{f, \mathcal{H}}(\mu ; P) + \epsilon \sup_{h \in \mathcal{H}} \Theta_{\mathcal{F}}(h),
\end{align*}
where $(1)$ holds since we can exchange supremums, $(2)$ is due to Theorem \ref{main-ipm-robustness-thm}, $(3)$ holds since $\Lambda_{\mathcal{F}, \epsilon} \leq \epsilon \Theta_{\mathcal{F}}(h)$ and finally $(4)$ holds since we can upper bound by taking out supremums.
\end{proof}
\newpage
\begin{lemma}\label{mu-ipm-der}
For any $\mu \in \mathscr{P}(\Omega)$, $h \in \mathscr{F}(\Omega, \mathbb{R})$ we have
\begin{align*}
    \inf_{b \in \mathbb{R}} \sqrt{\E_{\mu(X)}[\bracket{h(X)-b}^2] } = \sqrt{\operatorname{Var}_{\mu}(h) }
\end{align*}
\end{lemma}
\begin{proof}
Let $\varphi(b) = \E_{\mu(X)}[\bracket{h(X)-b}^2] $ and $S(b) = \sqrt{\varphi(b)}$ and using simple calculus we have
\begin{align*}
    S'(b) = \frac{\varphi'(b)}{2\sqrt{\varphi(b)}},
\end{align*}
and noting that $\varphi(b) > 0$, we can find the minima by solving $\varphi'(b) = 0$ by first noting that 
\begin{align*}
    \varphi(b) = \E_{\mu(X)}[h^2(X)] - 2b \E_{\mu(X)}[h(X)] + b^2,
\end{align*}
and so we have
\begin{align*}
    \varphi'(b) = 0 &\iff -2 \cdot \E_{\mu(X)}[h(X)] + 2b = 0\\
                    &\iff b =  \E_{\mu(X)}[h(X)].
\end{align*}
Putting this together yields
\begin{align*}
    \inf_{b \in \mathbb{R}} \sqrt{\E_{\mu(X)}[\bracket{h(X)-b}^2] } &=  \inf_{b \in \mathbb{R}} S(b) \\
    &= S\bracket{\E_{\mu(X)}[h(X)]}\\
    &= \E_{\mu(X)} \left[\bracket{h(X) - \E_{\mu(X)}[h(X)]}^2 \right]\\
    &= \sqrt{\operatorname{Var}_{\mu}(h) }
\end{align*}
\end{proof}

\end{document}